\def\year{2021}\relax
\documentclass[letterpaper]{article} 
\usepackage{aaai21}  
\usepackage{times}  
\usepackage{helvet} 
\usepackage{courier}  
\usepackage[hyphens]{url}  
\usepackage{graphicx} 
\usepackage{comment}
\usepackage{todonotes}
\usepackage{tikz}
\usepackage[title]{appendix}
\usetikzlibrary{shapes.geometric, arrows}
\usepackage{natbib}  
\usepackage{caption} 
\usepackage[utf8]{inputenc} 
\usepackage[T1]{fontenc}    
\usepackage{hyperref}       
\usepackage{url}            
\usepackage{booktabs}       
\usepackage{amsfonts}       
\usepackage{nicefrac}       
\usepackage{microtype}      
\usepackage{floatrow}
\usepackage{amssymb}
\usepackage{dblfloatfix}    
\usepackage{subcaption}
\usepackage{amsthm}
\usepackage[ruled]{algorithm2e}
\usepackage{mathrsfs}
\usepackage[inline]{enumitem}
\usepackage{calc}
\usepackage{amsmath, nccmath}

\usepackage{xcolor}

\newfloatcommand{capbtabbox}{table}[][\FBwidth]

\usepackage{amsmath,amsfonts,bm}

\def\eqref#1{equation~\ref{#1}}

\def\1{\bm{1}}

\DeclareMathAlphabet{\mathsfit}{\encodingdefault}{\sfdefault}{m}{sl}
\SetMathAlphabet{\mathsfit}{bold}{\encodingdefault}{\sfdefault}{bx}{n}

\newcommand{\E}{\mathbb{E}}

\DeclareMathOperator*{\argmax}{arg\,max}

\makeatletter
\providecommand*{\barvee}{%
  \mathbin{%
    \mathpalette\@barvee{}%
  }%
}
\newcommand*{\@barvee}[2]{%
  \sbox0{$#1\veebar\m@th$}%
  \sbox2{%
    \hbox to \wd0{%
      \hss
      \resizebox{1.05\wd0}{\height}{$#1-\m@th$}%
      \hss
    }%
  }%
  \sbox4{%
    \resizebox{\wd0}{.7\ht0}{$#1\vee\m@th$}%
  }%
  \sbox6{$#1\vcenter{}$}
  \ht2=\ht6 %
  \vbox to \ht0{%
    \copy2 %
    \vss
    \copy4 %
  }%
}
\makeatother

\newcommand{\state}{\mathcal{S}}
\newcommand{\action}{\mathcal{A}}
\newcommand{\dynamics}{p}
\newcommand{\reward}{r}
\newcommand{\rmax}{\reward_{\text{MAX}}}
\newcommand{\rmin}{\reward_{\text{MIN}}}
\newcommand{\goals}{\mathcal{G}}
\renewcommand{\v}{V}
\newcommand{\q}{Q}
\newcommand{\vpi}{V^\pi}
\newcommand{\qpi}{Q^\pi}
\newcommand{\vstar}{V^*}
\newcommand{\qstar}{Q^*}
\newcommand{\pistar}{\pi^*}

\usepackage{mathtools}

\newcommand{\rbar}{\bar{\reward}}
\newcommand{\qbar}{\bar{\q}}
\newcommand{\qstarbar}{\bar{\q}^*}

\newcommand{\qstarbarbig}{\qstarbar_{MAX}}
\newcommand{\qstarbarsmall}{\qstarbar_{MIN}}

\title{Learning to Follow Language Instructions with Compositional Policies

}
\author{
    Vanya Cohen,\textsuperscript{\rm 1$\dagger$\footnote{Work performed while an independent researcher.}} Geraud Nangue Tasse,\textsuperscript{\rm 2$\dagger$} Nakul Gopalan,\textsuperscript{\rm 3} \\ Steven James,\textsuperscript{\rm 2} Matthew Gombolay,\textsuperscript{\rm 3} Benjamin Rosman\textsuperscript{\rm 2}
    \\
}

\affiliations{
    \textsuperscript{\rm 1}The University of Texas at Austin\\
    \textsuperscript{\rm 2}University of the Witwatersrand\\
    \textsuperscript{\rm 3}Georgia Institute of Technology\\
    \textsuperscript{$\dagger$} Equal contribution\\
    vanya@utexas.edu,
    geraudnt@gmail.com,
    nakul\_gopalan@gatech.edu,
    steven.james@wits.ac.za,
    matthew.gombolay@cc.gatech.edu,
    benjamin.rosman1@wits.ac.za

}

\begin{document}

\maketitle

\begin{abstract}

We propose a framework that learns to execute natural language instructions in an environment consisting of goal-reaching tasks that share components of their task descriptions. 
Our approach leverages the compositionality of both value functions and language, with the aim of reducing the sample complexity of learning novel tasks. 
First, we train a reinforcement learning agent to learn value functions that can be subsequently composed through a Boolean algebra to solve novel tasks.
Second, we fine-tune a seq2seq model pretrained on web-scale corpora to map language to logical expressions that specify the required value function compositions.
Evaluating our agent in the BabyAI domain, we observe a decrease of $86\%$ in the number of training steps needed to learn a second task after mastering a single task.
Results from ablation studies further indicate that it is the \emph{combination} of compositional value functions and language representations that allows the agent to quickly generalize to new tasks.

\end{abstract}

\section{Introduction}
Natural language provides an intuitive way for people to specify tasks and instructions to satisfy goals. 
However, instruction following is a difficult problem for artificial agents because they need to simultaneously learn
the \begin{enumerate*}[label=(\alph*)]
  \item  meaning of the instructions to solve the task,
  \item  representation of the world in which the task is to be solved, and
  \item  sequence of actions that will lead to task completion.
\end{enumerate*}
Learning to follow instructions is even more challenging in the multitask setting, where an agent must also acquire knowledge that will allow it to generalize to new, unseen tasks. 

One of the most common approaches to this problem is to encode a language command into a real-valued vector embedding. 
The language embedding, together with the agent's current state, is then used to parametrize the policy that controls the agent~\cite{blukis2019learning,chaplot2018gated,tambwekar2021interpretable}. 
These end-to-end methods succeed in the presence of a simulator, but require large amounts of data. These methods also suffer from generalization issues as the agents require a large number of samples from the environment for every novel task presented~\cite{lake2018generalization}.

To overcome the generalization issue, we leverage the principle of \textit{compositionality}, which states that a complex expression's meaning is defined by the meanings of its constituent expressions and the rules used to combine the meanings of these expressions~\cite{sep-compositionality}.
In particular,  we exploit the compositional nature of task specifications and their solutions (in the form of a special type of value function). 
Our approach is possible as both natural language and the learned value functions exhibit the property of compositionality.

In our work, the constituent expressions are possible atomic goal specifications. These goal specifications have \emph{value functions} which define the agent's behavior. The value functions can be combined using the rules of Boolean algebra to solve novel tasks. We use machine translation approaches to map linguistic task specifications onto corresponding logical expressions, which are then used to combine value functions to solve the specified tasks.

First, we train task-specific value functions in the manner of~\citet{tasse2020boolean} for a set of preselected atomic tasks from the BabyAI environment~\cite{babyai_iclr19}. 
Additionally, we fine-tune a T5 model~\cite{raffel2019exploring} using reinforcement learning to translate natural language instructions into logical expressions that specify the compositions of the task-specific value functions. The compositional value functions are then used by the agent to form policies for acting in the environment. 
Finally, the agent's collected environment rewards are used as a signal to improve the translation model.

We evaluate the agent by learning a set of compositional tasks in series and observe the number of training steps needed to learn each additional task in the series. 
Further, we perform ablation studies to understand the effect of model pretraining on web-scale corpora and the stochastic nature of feedback from the environment on sample complexity. With a pretrained T5 model~\cite{raffel2019exploring}, the mean number of training steps needed to learn an additional task drops by $86\%$  after learning just one task. 
Without model pretraining, the mean training steps drops by only $6\%$, although the number of training steps continues to drop as more tasks are learned.

When learning all available tasks in the environment, the number of training steps needed to learn the final task decreases by $98\%$ for the pretrained model, compared to only $80\%$ for the randomly initialized model. In terms of the fractional improvement in the training steps needed to learn the final task, the pretrained model provides a $10\times$ improvement ($2\%$ versus $20\%$) over the randomly-initialized model.

Overall, our paper makes the following contributions:
\begin{itemize}
  \item We connect pretrained compositional policies to a translation model capable of mapping natural language statements to logical expressions specifying compositions of those policies.
  \item We demonstrate empirically that pretraining of the translation model on non-task-specific data is sufficient to generate compositional expressions. We found the T5~\cite{raffel2019exploring} language model sufficient to generate novel Boolean expressions given language commands. This ability to generate novel expressions leads to a significant reduction in samples from the environment after the pretrained model learns to solve just a single task.
  \item We detail learning results for $18$ tasks in the BabyAI showing that compositional policies, along with a pretrained model, lead to substantial savings in the number of samples required to learn novel tasks.
  \item We provide ablation results with and without a pretrained language model, and with and without our compositional policies evaluated in the environment.
\end{itemize}

\begin{figure*}[h!]
    \centering
    \begin{subfigure}[t]{0.3\textwidth}
        \centering
        \includegraphics[width=1.5in]{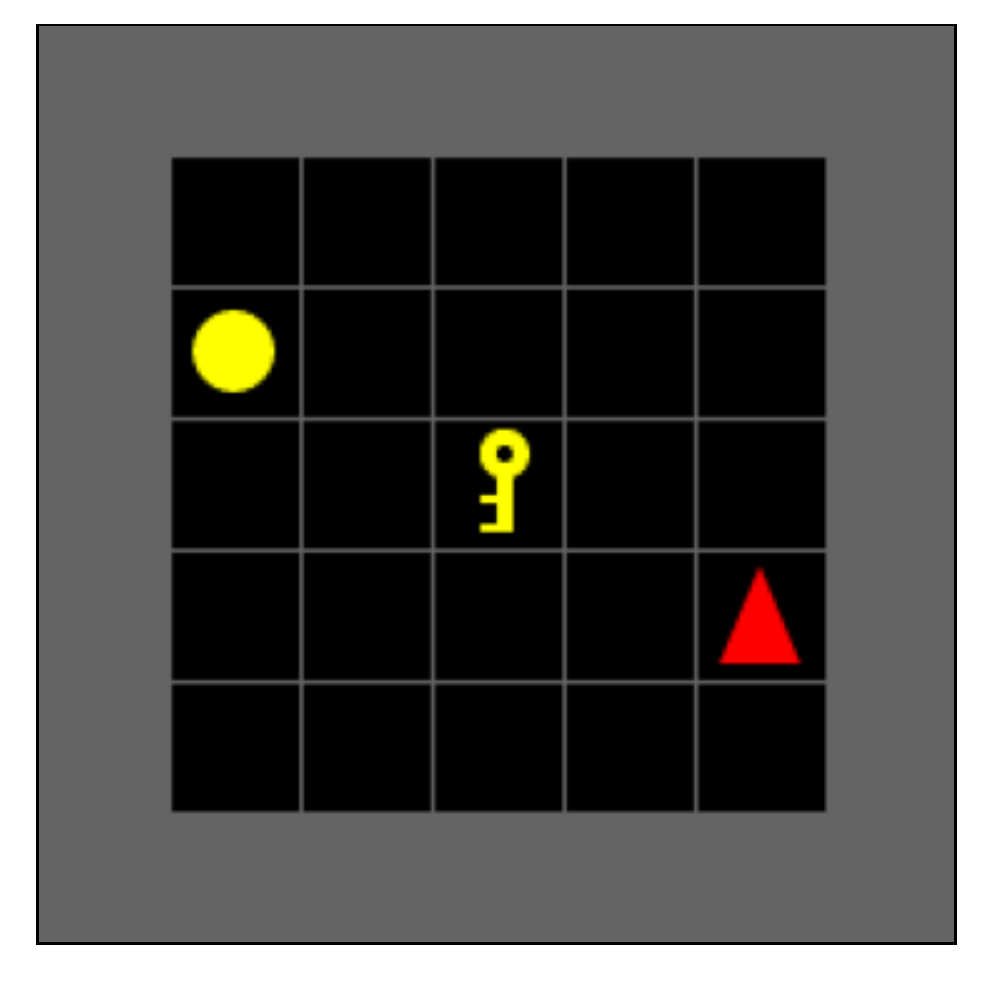}
        \caption{Language command ``pick up the yellow ball'' with corresponding logical expression \texttt{pickup\_yellow} $\wedge$ \texttt{pickup\_ball}.}     \label{fig:enva}
    \end{subfigure}%
    \quad
    \begin{subfigure}[t]{0.3\textwidth}
        \centering
        \includegraphics[width=1.5in]{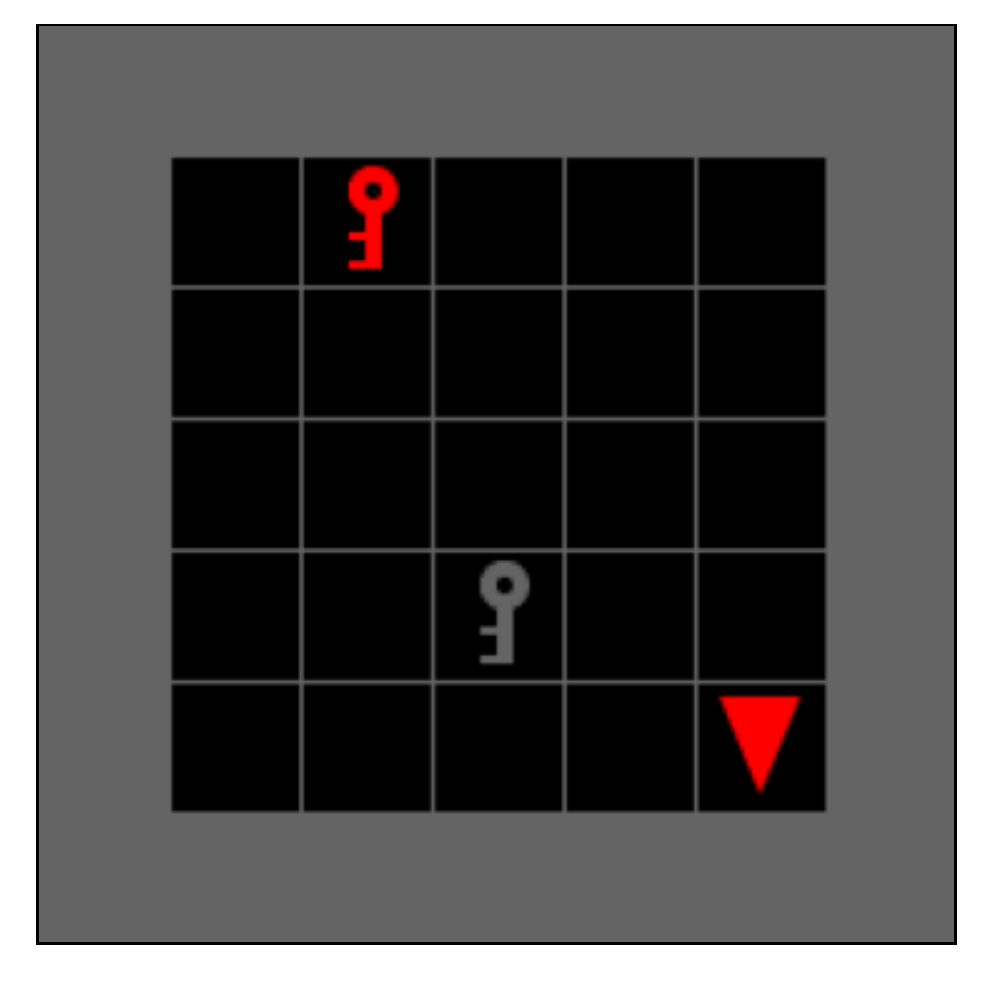}
        \caption{Language command ``pick up the red key'' with corresponding logical expression \texttt{pickup\_red} $\wedge$ \texttt{pickup\_key}.}     \label{fig:envb}

    \end{subfigure}%
    \quad
    \begin{subfigure}[t]{0.3\textwidth}
        \centering
        \includegraphics[width=1.5in]{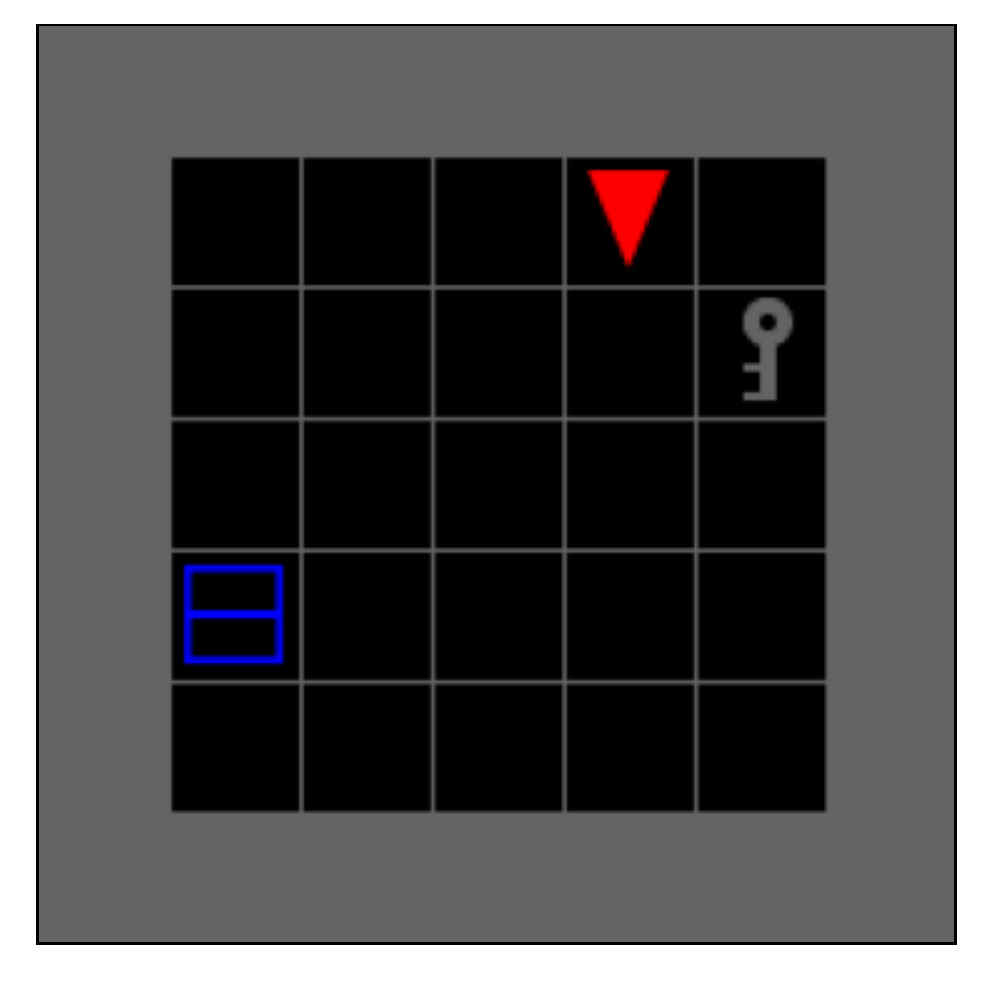}
        \caption{Language command ``pick up the blue box'' with corresponding logical expression \texttt{pickup\_blue} $\wedge$ \texttt{pickup\_box}.}    \label{fig:envc}

    \end{subfigure}%
    \caption[ ]{
    Examples of tasks in the BabyAI \texttt{PickUpObj} environment. For each task, there is a target and distractor object. The agent is represented by the red triangle. We also investigate performance when four distrator objects are present. \subref{fig:enva}) The agent must pick up the yellow ball but not the yellow key. To solve this level, the agent must use the intersection of the ``pickup'' value functions for ``yellow'' and ``ball''. \subref{fig:envb}) The agent must pick up the red key while not picking up the grey key. Solving this level requires using the intersection of the ``pickup'' value functions for ``red'' and ``key''. \subref{fig:envc}) The agent must pick up the blue box while avoiding picking up the grey key. 
    }
    \label{fig:env}
\end{figure*}

\section{Background}

We are interested in the multitask setting, where an agent is required to solve a series of related tasks, modelled by a unordered set of Markov Decision Processes (MDPs).
An MDP is defined by the tuple $\langle\mathcal{\state}, \action, \dynamics, \reward\rangle$, where 
\begin{enumerate*}[label=(\roman*)]
  \item $\state$ is the state space,
  \item $\action$ is the action space,
  \item $\dynamics$ is a Markov transition kernel $(s, a) \mapsto \rho_{(s, a)}$ from $\state \times \action$ to $\state$, and
  \item $\reward$ is the real-valued reward function bounded by $[\rmin, \rmax]$.
\end{enumerate*}
We focus here on stochastic shortest path problems~\citep{bertsekas91}, where an agent must optimally reach a set of absorbing goal states $\goals \subseteq \state$.

We assume that all tasks share the same state space, action space and dynamics, but differ in their reward functions. More specifically, we define the background MDP $M_0 = \langle \mathcal{\state}_0, \action_0, \dynamics_0, \reward_0 \rangle$ with its own state space, action space, transition dynamics and background reward function. 
Any individual task $\tau$ is specified by a task-specific reward function $\reward_\tau$ that is non-zero only for transitions entering a state in $\goals$.
The reward function for the resulting MDP is then simply $\reward_0 + \reward_\tau$.

Given a task, the agent's aim is to learn an optimal Markov policy $\pi$ from $\state$ to $\action$.
A given policy $\pi$ induces a value function $\vpi(s) = \E_\pi \left[ \sum_{t=0}^{\infty} r(s_t, a_t) \right]$, representing the expected return obtained under $\pi$ starting from state $s$.
The \textit{optimal} policy $\pistar$ is the policy that obtains the greatest expected return at each state: $\v^{\pistar}(s) = \vstar(s) = \max_\pi \vpi (s)$ for all $s \in \state$.
A related quantity is the $Q$-value function, $\qpi(s, a)$, which defines the expected return obtained by executing $a$ from $s$, and thereafter following $\pi$.
Similarly, the optimal $Q$-value function is given by $\qstar(s, a) = \max_\pi \qpi(s, a)$ for all $(s, a) \in \state \times \action$.
Finally, we define a \textit{proper policy} to be a policy that is guaranteed to eventually reach $\goals$ \citep{james06,vanniekerk19}. 
We assume the value functions for improper policies---those that never reach absorbing states---are unbounded from below.

\subsection{Logical Composition of Tasks and Value Functions}

\citet{tasse2020boolean} recently proposed a framework for agents to apply logical operations---conjunction ($\wedge$), disjunction ($\vee$) and negation ($\neg$)---over the space of tasks and value functions.
This is achieved by first defining the goal-oriented reward function $\rbar$ which extends the task rewards $r$ to penalise an agent for achieving goals different from the one it wished to achieve:
\begin{equation}
    \rbar(s, g, a) = \begin{cases}
\bar r_{MIN} & \text{if } g \neq s \in \goals\\
r(s, a) &\text{otherwise},
\end{cases}
\label{eq:erf}
\end{equation}
where $\rbar_{MIN} \leq \min \{\rmin, (\rmin - \rmax)D\}$, and $D$ is the diameter of the MDP \citep{jaksch10}.
Equation~\ref{eq:erf} specifies the reward function for the agent to achieve all reachable goals. 

Using Equation~\ref{eq:erf}, we can define the related goal-oriented value function as:
\begin{equation}
    \qbar(s, g, a) = \rbar(s, g, a) + \int_{\state} \bar{\v}^{\bar{\pi}}(s^\prime, g) \dynamics_{(s, a)} (ds^\prime),
\end{equation}
where $\bar{\v}^{\bar{\pi}}(s, g) = \E_{\bar{\pi}} \left[ \sum_{t=0}^{\infty} \rbar(s_t, g, a_t) \right]$.

If a new task can be represented as the logical expression of previously learned tasks,~\citet{tasse2020boolean} prove that the optimal policy can immediately be obtained by composing the learned goal-oriented value functions using the same expression.

For example, consider the \texttt{PickUpObj} domain shown in Figure~\ref{fig:env}, where an agent has learned to pick up the yellow object (task $Y$) and to pick up the ball (task $B$). 
We can then provably solve the tasks defined by their union, intersection, and negation as follows (we omit the value functions' parameters for readability):

\begin{alignat*}{3}
  \qstarbar_{Y \vee B}  = \qstarbar_{Y} \vee \qstarbar_{B} \coloneqq  \max\{\qstarbar_{Y}, \qstarbar_{B}\}
\end{alignat*} 
\begin{alignat*}{3}
  \qstarbar_{Y \wedge B} = \qstarbar_{Y} \wedge \qstarbar_{B} \coloneqq \min\{\qstarbar_{Y}, \qstarbar_{B}\}
\end{alignat*} 
\begin{alignat*}{3}
  \qstarbar_{\neg Y} = \neg \qstarbar_{Y} \coloneqq \left(\qstarbarbig + \qstarbarsmall \right) - \qstarbar_{Y},
\end{alignat*} 
where $\qstarbarbig$ is the goal-oriented value function for the maximum task where $r = \rmax$ for all $\goals$.
Similarly $\qstarbarsmall$ is goal-oriented value function  for the minimum task where $r = \rmin$ for all $\goals$. 
We henceforth refer to these goal-oriented value functions as \textit{compositional value functions}.

\subsection{Translation with Transformer Models}

Recent progress in natural language processing (NLP) has demonstrated the effectiveness of large-scale generative pretraining and subsequent fine-tuning on downstream tasks, such as translation, question answering, and classification \citep{devlin2018bert,peters2018deep,radford2018improving}. Subsequent work has shown that scaling both model parameters and pretraining corpus size leads to better transfer learning and generalization \cite{radford2019language}. 

To map between natural language instructions and Boolean expressions specifying policy compositions, we utilize the T5 sequence-to-sequence model \cite{raffel2019exploring} based on the Transformer architecture \cite{vaswani2017attention}. The model is pretrained using an unsupervised learning objective on the Colossal Clean Crawled Corpus (C4) \cite{raffel2019exploring}, a filtered version of the Common Crawl.\footnote{https://commoncrawl.org} The C4 corpus contains 750GB of text, the vast majority of which is fluent English. \citeauthor{raffel2019exploring} perform exhaustive ablation studies to develop their pretrained models, which offer good performance on a variety of NLP tasks including translation.

Transformer-based models use the self-attention mechanism \cite{vaswani2017attention} to build sequence representations of text inputs, and to transform those representations into probability distributions over text outputs. As with the original Transformer architecture, the T5 model is composed of both an encoder and decoder stack of self-attention layers to map input sequences to output sequences. Self-attention layers receive input embeddings from lower layers and compose them to form higher-level embeddings.

\section{Methods}

Our agent learns to combine pretrained compositional policies by translating BabyAI ``mission'' statements (e.g. ``pick up the blue box'') into Boolean algebraic expressions which specify compositions of policies. We limit our investigation to intersections of policies, although the Boolean compositional policies also allow for disjunction and negation. Training begins with training a compositional policy to solve each of the task primitives. The agent can navigate to objects in the BabyAI domain described by three type attributes $\{box, ball, key\}$ and six color attributes $\{red,\linebreak[0] blue,\linebreak[0] green,\linebreak[0] grey,\linebreak[0] purple,\linebreak[0] yellow\}$, which yields eighteen possible navigation tasks.

\subsection{Learning the Compositional Value Functions}

Like \citet{tasse2020boolean}, we use deep Q-learning \cite{mnih15} to learn the Q-function for each goal of the compositional value functions. 
We represent each compositional value function $\qbar^*$ with a list of $|\goals|$ DQNs, such that the Q-function for each goal $Q_g^*(s,a) \coloneqq \qbar^*(s,g,a)$ is approximated with a separate DQN.\footnote{Architecture and hyperparameter details are presented in the appendix.}

For each task, the agent starts training after 1000 steps of random exploration to populate an experience replay buffer and a goal buffer (set of reached terminal states). For each episode, the agent samples a random goal from the goal buffer and uses $\epsilon$-greedy to act in the environment. For each action, $a$, that the agent takes in each state, $s$, it receives goal-oriented rewards (Equation \ref{eq:erf}) given by:
\begin{equation*}
    \rbar(s, g, a) = \begin{cases}
 -0.1 & \text{if } g \neq s \in \goals\\
r(s, a) &\text{otherwise},
\end{cases}
\end{equation*}
where task reward $r(s,a)=2$ for picking up the correct object and $r(s,a)=-0.1$ everywhere else.\footnote{We used $\rbar_{MIN}=r_{MIN}=-0.1$ since that is the simplest choice and it did not result in any discernible change in the success rate of the composed policies.} The episode terminates after the agent picks up any object.
The agent's compositional value function is then trained per episode using the collected experience. Training ends once the agent reaches a success rate of at least $0.98$. For lower success rates, the compounding effect of composing sub-optimal policies negatively impacts the translation model's learning. For full details on policy performance please refer to Table \ref{policyperformance} in Appendix A.

\subsection{Translating Missions to Boolean Expressions}

We select the smallest of the publicly released T5 models as the pretrained model for our experiments (Table~\ref{t5params}): the T5-small model which has 60 million parameters and is sufficient for our tasks based upon our empirical exploration. 

We translate natural language task instructions to Boolean algebraic expressions that represent the task's value function. 
The Boolean algebraic expressions have tokens and operators. The legal operators are union (disjunction), intersection (conjunction) and negation. 
The tokens in the Boolean algebraic expressions represent goal value functions that can be composed to create richer tasks. For the BabyAI domain, these tokens represent value functions for picking up objects by type $\{pickup\_box,\linebreak[0] pickup\_ball,\linebreak[0] pickup\_key\}$, picking up objects by color $\{pickup\_red,\linebreak[0] pickup\_blue,\linebreak[0] pickup\_green,\linebreak[0] pickup\_grey,\linebreak[0] pickup\_purple,\linebreak[0] pickup\_yellow\}$, the logical operators and end-of-sentence tokens $\{and,\linebreak[0] <s>\}$.

\begin{figure}[tb!]
\vskip 0.2in
\begin{center}
\centerline{\includegraphics[width=0.9\columnwidth]{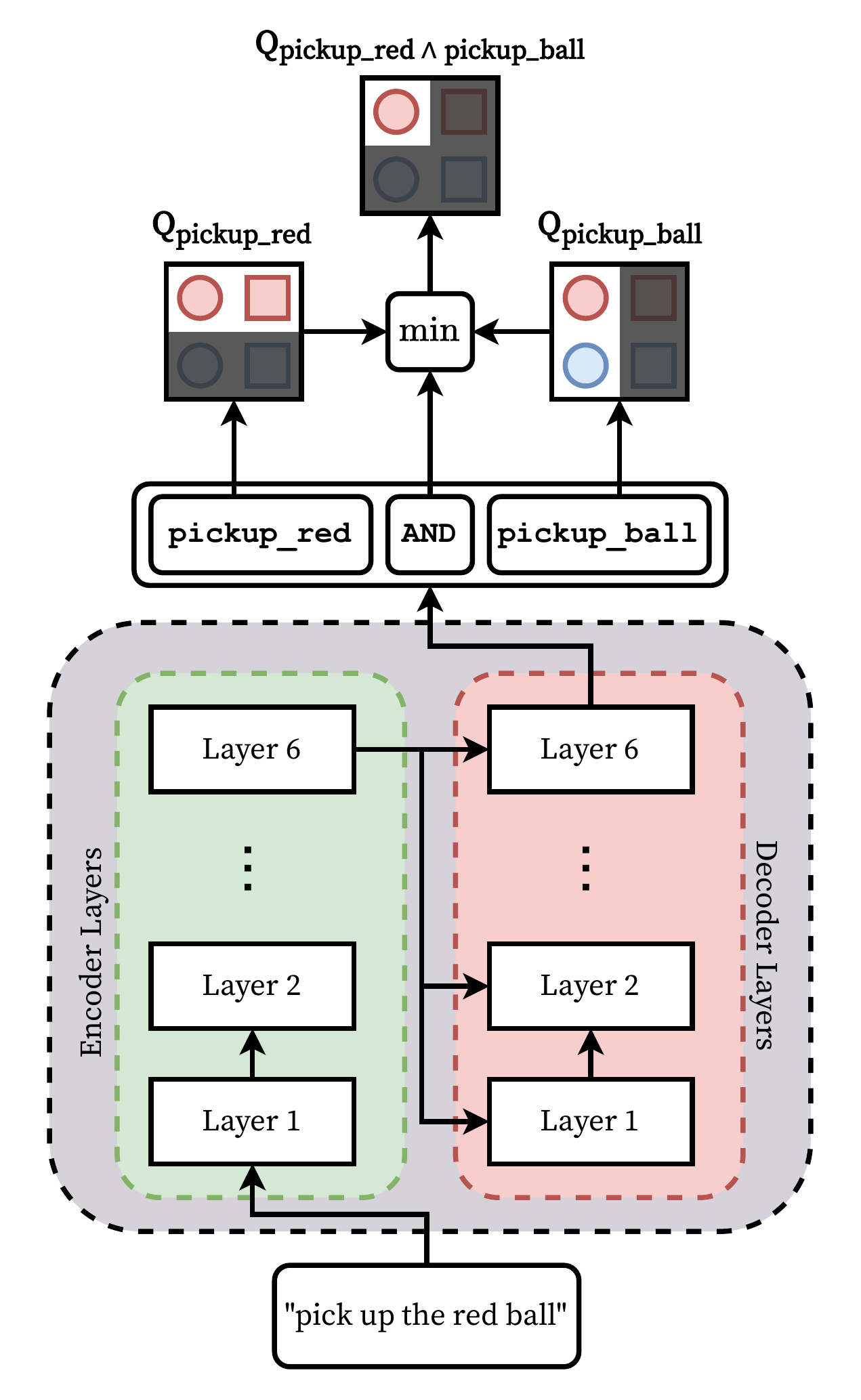}}
\caption{The T5-small model first translates the input mission command ``pick up the red ball'' into a Boolean expression, with variables representing the vocabulary of possible compositional value functions. Then the intersection of the value functions is computed, resulting in a value function for picking up a red ball in the environment.}
\label{fig:process}
\end{center}
\vskip -0.2in
\end{figure}

Both the input and output tokens are byte-pair encoding (BPE) subword units \cite{sennrich2016neural} learned from the C4 training corpus. For example, the Boolean task algebra token $``pickup\_purple''$ is represented by the subwords $``pickup''$, $``\_''$, $``pur''$, and $``ple''$. Each of the tokens in the Boolean task algebra is represented by one or more BPE subword units. Instead of sampling from the BPE subword units directly, continuations are sampled from the distribution of tokens in the Boolean task algebra. If BPE subwords were sampled directly, at the beginning of training the probability of outputting valid tokens from the Boolean task algebra would be vanishingly small. Decoding stops when the stop token is produced or more than three Boolean algebra tokens have been sampled. We use temperature-based sampling \cite{ackley1985learning} to produce translated sequences during training, and greedy sampling during evaluation. For translation model details see Table \ref{t5params}.

Given an input mission to the T5-small model (e.g. ``pick up the red ball'') we can sample a Boolean expression from its output distribution over tokens (e.g. \texttt{pickup\_red and pickup\_ball}). This expression is then parsed and validated for syntactic correctness by a Boolean algebra expression parser. The corresponding compositional value function is obtained as follows (we omit the value functions' parameters for readability):

\begin{alignat*}{3}
  \qbar_{\textit{pickup\_red} \wedge \textit{pickup\_ball}} = \min\{\qbar_{\textit{pickup\_red}}, \qbar_{\textit{pickup\_ball}}\}
\end{alignat*} 

The full process for generating policies from task instructions is illustrated in Figure~\ref{fig:process}. Finally, the agent can maximise over the composed value function to act in the environment: $\pi(s) \in \argmax_{a \in \action} \max_{g \in \goals} \qbar(s, g, a)$.

\subsection{Baseline Model}
The baseline is a non-compositional CNN-DQN \cite{mnih15} conditioned on the input mission language. The model is a simplified version of the baseline used by \citet{babyai_iclr19}, and uses a CNN to extract image features and a Gated Recurrent Unit (GRU) \cite{cho2014learning} that takes the mission as input and outputs text features. The image and text features are then concatenated and passed through two fully-connected layers to compute the output Q-values.

In contrast to our method, the baseline is a joint model which learns a Q-function conditioned on both image state and language features. As such, its component value function and language representations are not pretrained. Our method learns both tasks and language separately and then learns to combine them compositionally. Likewise, the baseline model does not have explicit compositional structure and must instead learn to condition output Q-values on a combination of image and language features.

\section{Experiments}

To assess the agent's ability to generalize compositionally, we evaluate the agent as it learns to solve all available tasks in sequence. In each of ten trials, we randomly shuffle the order in which the $18$ tasks are introduced and then train the agent to solve each task one at a time.
At iteration $0$ of each new task, and every $100$ training steps thereafter, the performance of the agent is evaluated using returns from $100$ policy roll-outs.
A task is considered solved if the agent successfully reaches the goal object in $95$ out of $100$ roll-outs, at which point the agent is presented with the next task in the sequence.
During training, Boolean policy expressions are sampled from the translation model using temperature-based sampling with a temperature of $1.0$ to inject randomness in the sampling process. 
However, when evaluating whether the agent has successfully solved the task, expressions are generated through greedy sampling to only select the most likely continuation tokens without noise.

\subsection{Learning Tasks in Series}

\begin{figure*}[h!]
    \centering
    \begin{subfigure}[t]{0.46\textwidth}
        \centering
        \includegraphics[width=0.98\textwidth]{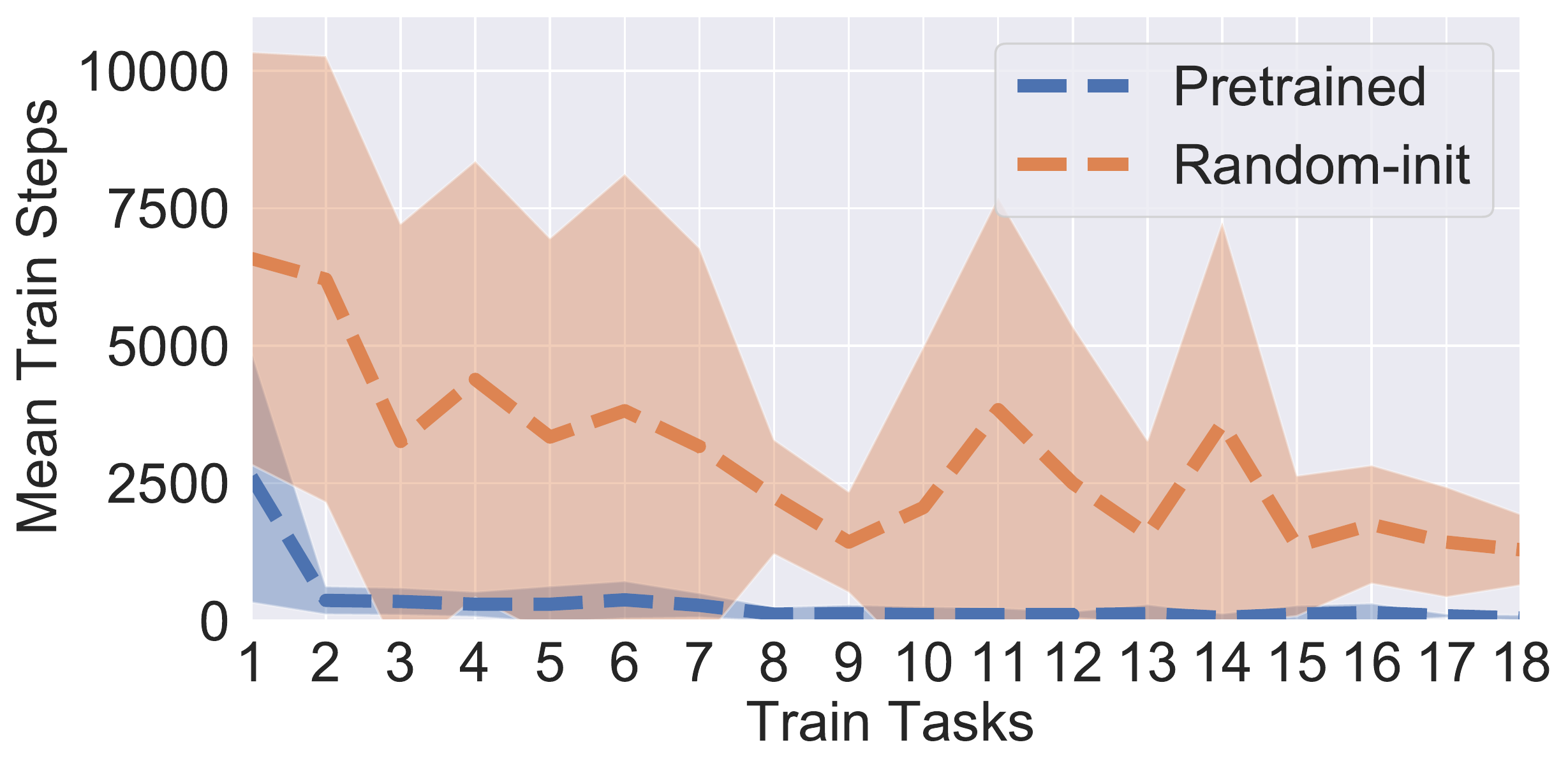}
        \caption{With feedback from evaluating the compositional policies in the environment with one distractor, compares the effect of pretraining on the number of training steps needed to learn each translation from each mission to each Boolean expression.}
        \label{fig:resa}
    \end{subfigure}%
    \quad
    \begin{subfigure}[t]{0.46\textwidth}
        \centering
        \includegraphics[width=0.98\textwidth]{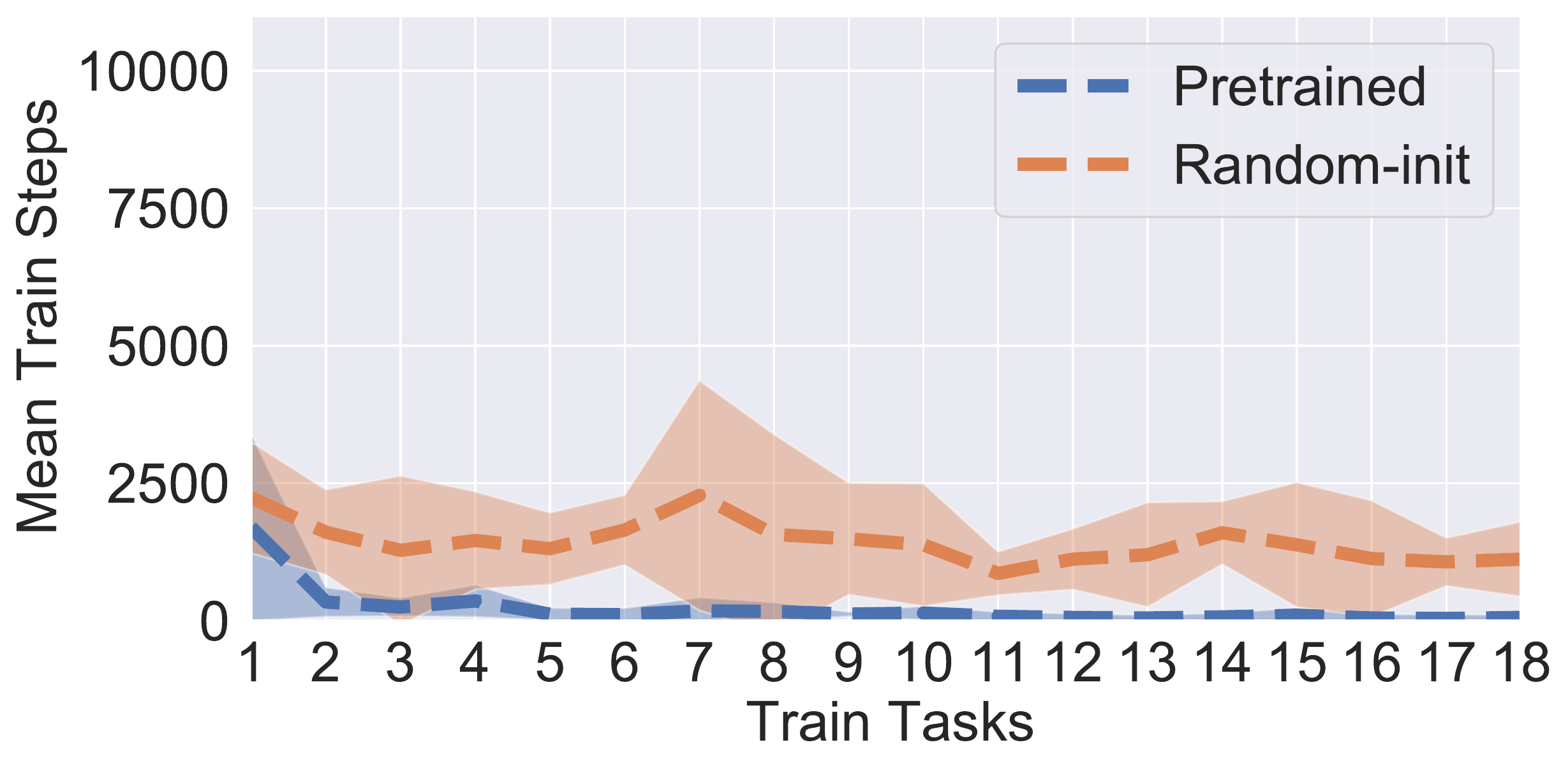}
        \caption{With ``perfect'' feedback based only on the output Boolean expression, compares the effect of pretraining on the number of training steps needed to learn each translation from each mission to each Boolean expression.
        }
        \label{fig:resb}
    \end{subfigure}%
    \par\medskip
    \begin{subfigure}[t]{0.46\textwidth}
        \centering
        \includegraphics[width=0.98\textwidth]{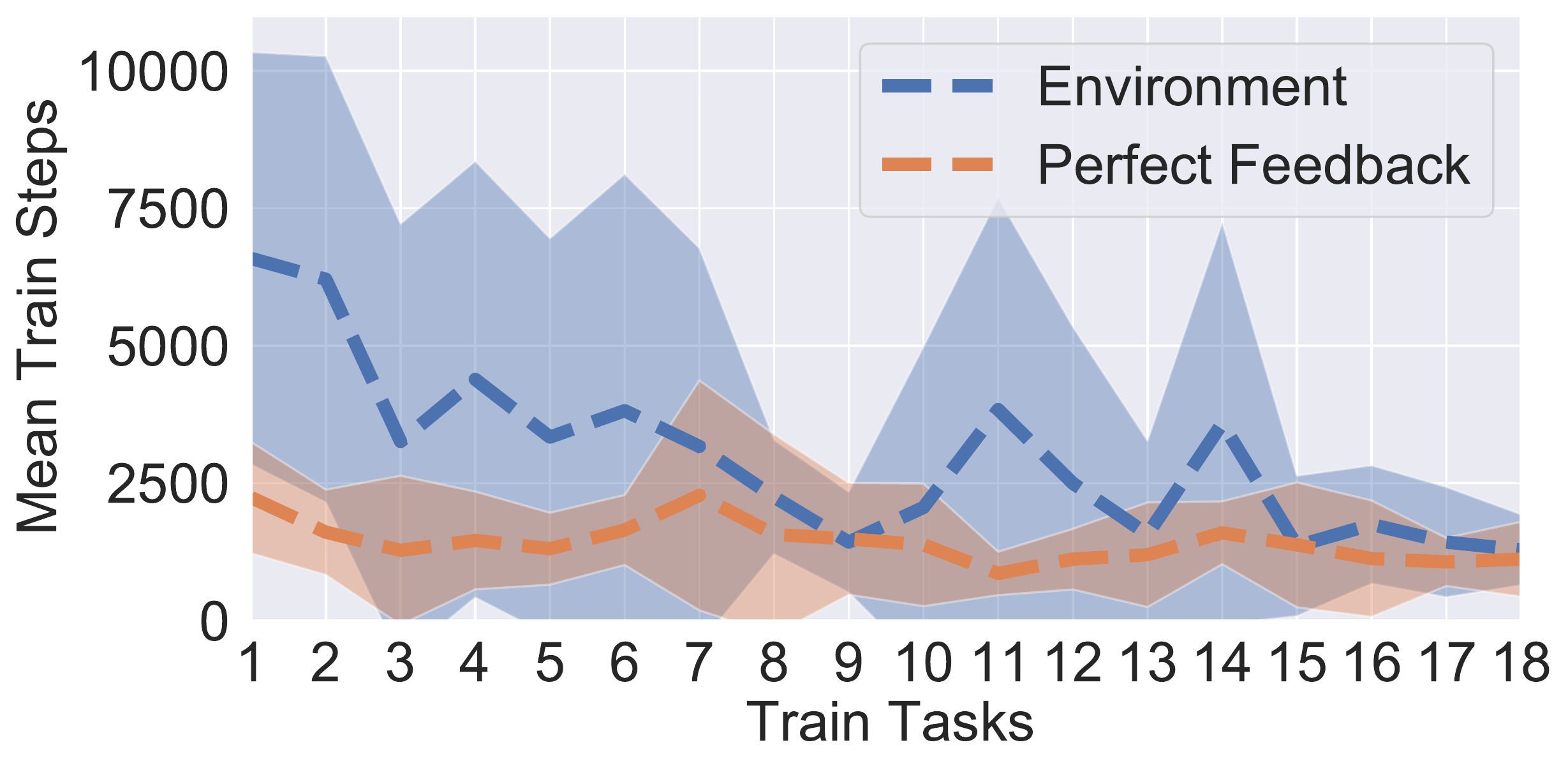}
        \caption{With a randomly-initialized T5 model, compares the learning of the agent from policies evaluated in the environment with one distractor to ``perfect'' feedback based only on the output Boolean expression. 
        }
        \label{fig:resc}
    \end{subfigure}
    \quad
    \begin{subfigure}[t]{0.46\textwidth}
        \centering
        \includegraphics[width=0.98\textwidth]{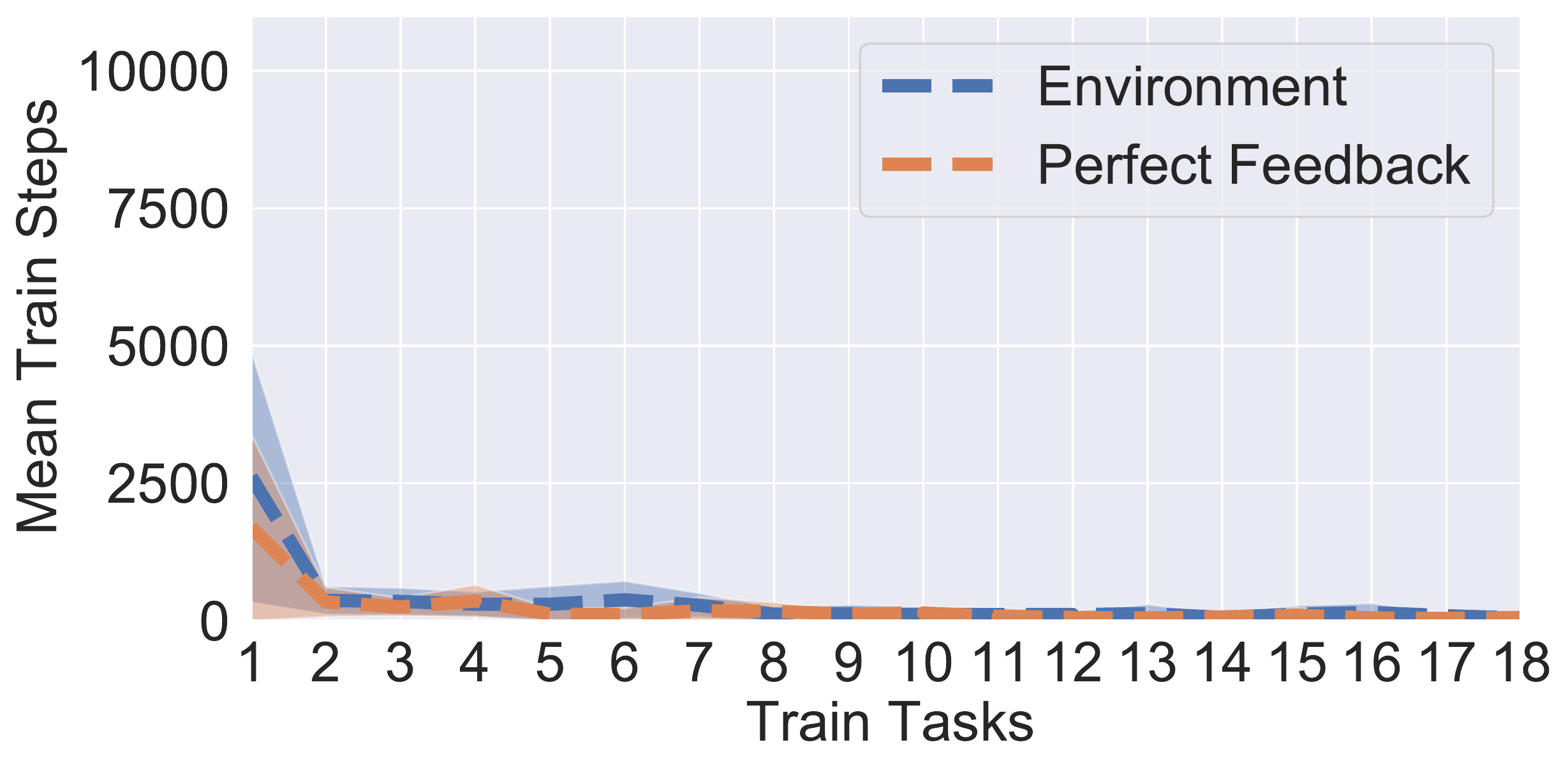}
        \caption{With a pretrained T5 model, compares the learning of the agent from policies evaluated in the environment with one distractor to ``perfect'' feedback based only on the output Boolean expression. 
        }
        \label{fig:resd}
    \end{subfigure}
    \begin{subfigure}[t]{0.46\textwidth}
        \centering
        \includegraphics[width=0.98\textwidth]{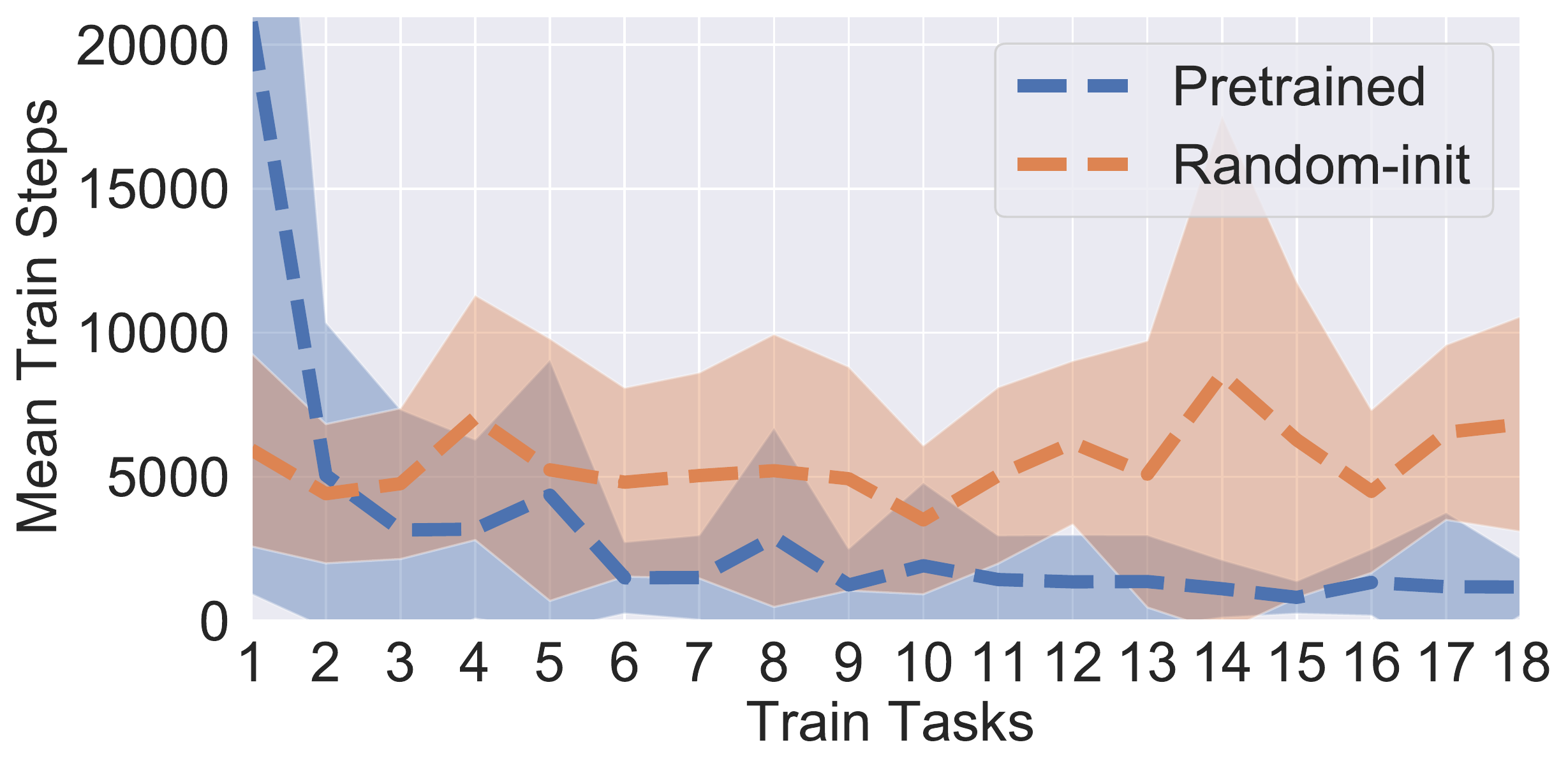}
        \caption{With four distractor objects and feedback from evaluating the compositional policies in the environment, compares the effect of pretraining on the number of training steps needed to learn each translation from each mission to each Boolean expression.
        }
        \label{fig:rese}
    \end{subfigure}
    \quad
    \begin{subfigure}[t]{0.46\textwidth}
        \centering
        \includegraphics[width=0.98\textwidth]{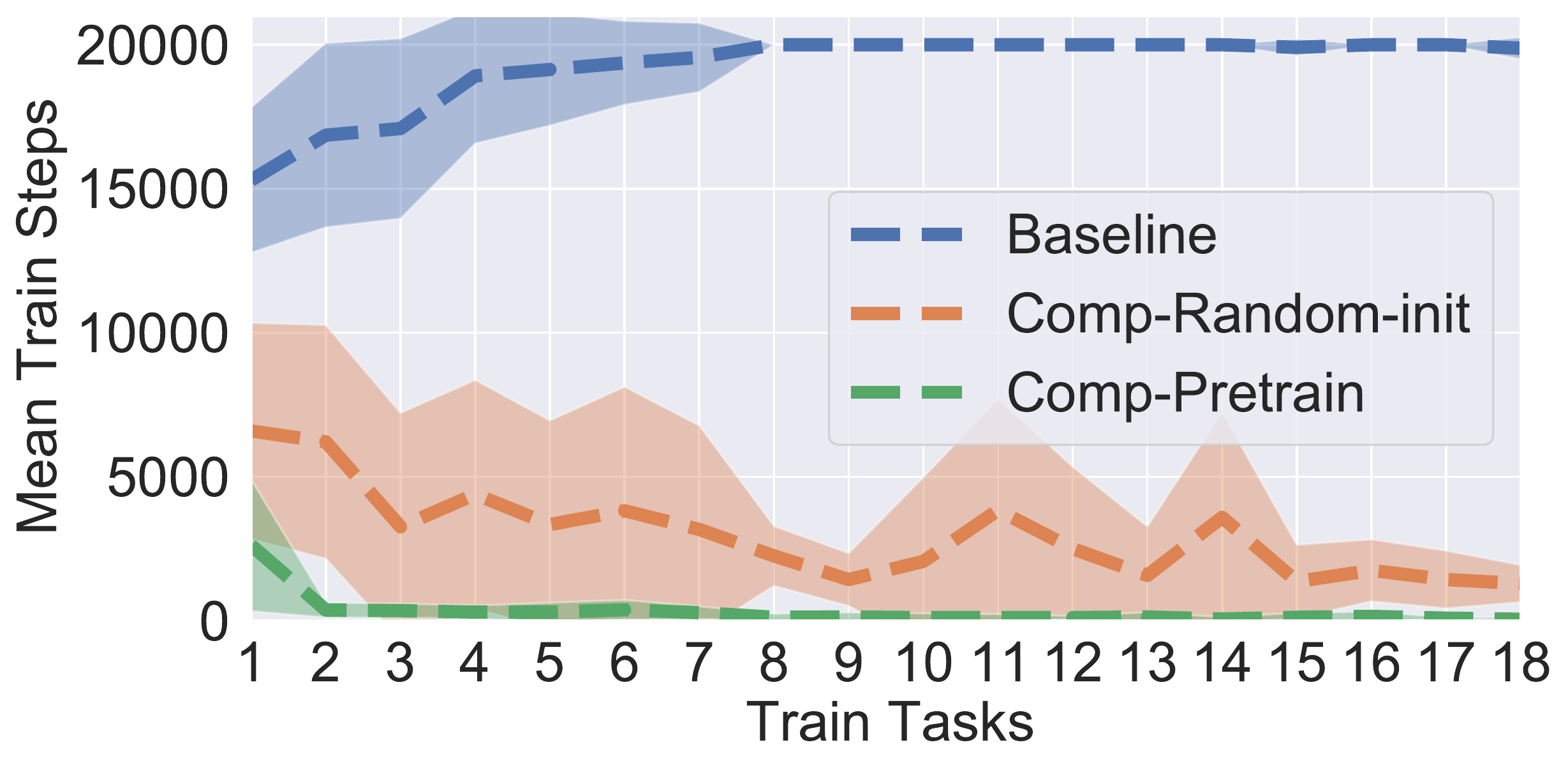}
        \caption{Performance of the Baseline model versus Compositional models (both with and without pretraining). All three models are trained using feedback from the environment with a single distractor object. While the baseline model requires substantially more training steps to learn each task, it does not poses explicit compositional structure and does not benefit from pretrained value functions or language representations.
        }
        \label{fig:resf}
    \end{subfigure}
    \caption[ ]{
    Number of training steps required by various agents to solve each task in a random sequence of tasks.
    The translation model used is the T5-small model with and without pretraining. Tasks are learned in series, with the same model used across tasks. Task order is randomized across trials. The shaded regions represent the standard deviations over 10 runs.
    }
    \label{fig:results1}
\end{figure*}

We adopt four experimental settings to investigate the impact of the different components in our overall system. 

First, we consider two strategies for initializing the translation model:  we use either 1) the pretrained T5-small model, or 2) its randomly initialized instantiation provided by \citet{wolf-etal-2020-transformers}.
We also consider the effect of the pretrained policies on the overall performance of the agent by 1) using the returns of the policies executed in the environment as a learning signal for the language model, or 2) directly comparing the output of the language model to the ground-truth logical expression. The combinations of language model pretraining and feedback type form the four experiments presented.

During training the environment provides noisy feedback from randomization of object and agent positions and imperfections in the trained compositional policies.
Further, each of the environments has one or four distractor objects sampled uniformly at random from the $18$ object types. 
These objects may have the same type attributes as the target object, in which case the mission command changes from using the definite to the indefinite article (e.g. ``pick up a red ball,'' instead of ``pick up the red ball'').

During inference the mission statement for the current task is translated to a Boolean expression, which is passed to a Boolean expression parser to determine syntactic correctness. 
If the expression is not syntactically valid, the agent receives a reward of $-1.0$.  
If the expression is valid, the corresponding compositional policy is instantiated and executed in the environment $50$ times.

The agent receives the mean reward from these $50$ roll-outs. Each episode receives a reward of $+1.0$ for successfully picking up the target object, and $-2.0$ for failing to pickup the target object within $50$ time-steps. The $50$ roll-out parameter was chosen by imperially establishing that $50$ offers a robust estimation of the mean reward for the instantiated policy.

The rewards of $+1.0$ and $-2.0$ were determined empirically to incentivize the production of optimal Boolean expressions. 
Without asymmetrically discouragement for picking up the wrong objects, the agent can learn to simply rely on color or object type (rather than both) to act in the environment and attain reward. 
This behavior represents a local minimum, where the agent will attain reward in cases where color or object type distinguishes the correct object from a distractor object. 
By asymmetrically discouraging failure, the agent is incentivized to utilize both the color and object type information to execute more precise policies.
The reward scale and sign determines whether the output Boolean expressions are made more or less likely by the cross-entropy loss.

Additionally, we evaluate the effect of environment noise on the translation model's learning. 
The translation model is separately trained using feedback from logically comparing sampled Boolean expressions to the known true Boolean expressions for those tasks. 
In this setting, the translation model receives a reward of $+1.0$ for outputting equivalent Boolean expressions and $-1.0$ for non-equivalent expressions. This removes sources of noise in training the language model: the environmental randomization, distractor objects, and noise from imperfect policies. 
However, it differs from purely supervised learning in that learning only occurs on samples from the translation model, produced through temperature sampling.

\begin{figure*}[h!]
\vskip 0.2in
\begin{center}
\centerline{\includegraphics[width=\textwidth]{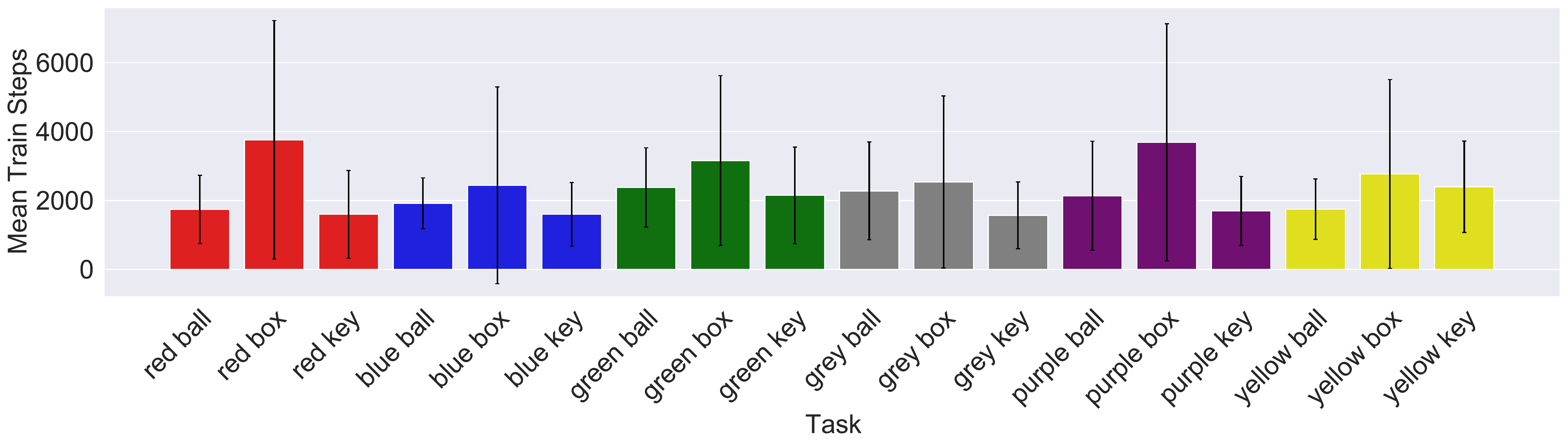}}
\caption{The mean training steps needed to learn the translation from each mission to each Boolean expression for each of the $18$ potential tasks, when the translation model has perfect feedback from the environment. 
Rewards are $+1.0$ for equivalent output Boolean expressions and $-1.0$ for incorrect expressions. Means and Standard Deviations computed over 10 trials.}
\label{lm-supervised}
\end{center}
\vskip -0.2in
\end{figure*}

Figure \ref{fig:results1} depicts the effects of pretraining versus randomly initializing the translation model, when training with feedback from the environment, and with feedback from the equivalence of output logical expressions. The results indicate that whether acting in the real environment or with ``perfect'' feedback based on logical equivalence, using the pretrained model vastly outperforms the randomly-initialized translation model. While both models see a decrease in the mean train steps across the randomly shuffled $18$ tasks, the number of samples required by the pretrained model drops precipitously after learning the first task. Further, in both the pretrained and randomly-initialized cases, learning in the environment is detrimental to model performance, with both the mean number of training steps, and the standard deviation higher for the agent when learning from environmental feedback. The greater number of training steps demonstrates the negative impact of the distractor objects and the agent's imperfect policies on translation model learning. In Figure \ref{fig:rese} the addition of more distractor objects initially requires more training steps on average to learn new tasks, but with substantially higher variance. However, as more tasks are learned, the pretrained model outperforms the randomly initialized model (as in the single distractor setup). We speculate that this is due to the higher variance in the rewards when using more distractor objects.

\subsection{Baseline Comparison}
Figure \ref{fig:resf} compares the number of training steps needed by the BabyAI baseline model to our compositional model. As with the other experiments, results are reported over 10 trials, where the agent learns to solve each task in the task set sequentially. The task order is shuffled between trials. In this experiment the number of training steps is capped at $20,000$ and a single distractor object is present in the environment. Initially, the baseline model succeeds in learning the first several tasks. However,  this model eventually begins to overfit, and reaches the training step limit for the remaining tasks in the set. Despite a much larger parameter count, neither of the compositional models overfit, and the compositional model with language model pretraining needs close to zero additional samples to learn the later tasks.

\subsection{Difficulty of Translation Tasks}

In this experiment, we fine-tune the pretrained translation model using reinforcement learning individually on each of the 18 tasks to compare the relative difficulty of the underlying translations. 
Unlike in the serial task learning experiment, the translation model learns each task individually with no transfer between tasks. The mean train steps and standard deviations are plotted for 10 trials for each task. The purpose is to determine if learning any of the translations for the tasks are significantly more challenging to learn than the others, which would lead to differential performance when learning certain sequences of tasks.

Figure \ref{lm-supervised} shows the mean train steps needed to learn each translation task based on the logical equivalence of the output expression to the ground-truth expression. The figure shows a similar range of difficulty in translating from each mission statement to each logical expression, indicating that no tasks are overwhelmingly more difficult than the others. However, there are differences in the translation difficulty between certain tasks. Translations for picking up ``box'' objects consistently require more training samples to learn. The unequal difficulty could be due to differences between the pretrained features for box objects.

\section{Related Work}

Instruction following by an artificial agent combines reasoning about an agent's current state, the language command given, and the best sequence of actions an agent can take to solve the given task, making it a challenging problem to solve.  
Previous approaches have attempted to solve this problem using a single deep neural network architecture trained using reinforcement learning~\cite{anderson2018vision,blukis2019learning,chaplot2018gated}. Another approach to solve this problem is to translate natural language into a sequence of symbols that can then be given as input to a planner~\cite{gopalan2018sequence}. It is possible to learn these symbols directly from data~\cite{gopalan2020simultaneously}, and compose these symbols using semantic parsing~\citep{dzifcak2009and,williams2018learning} to create novel task specifications that can then be planned over.

Previous work has also attempted to learn compositional linguistic representations to solve instruction following tasks~\cite{kuo2021compositional}. We, however, do not attempt to train and use compositional language representations~\cite{andreas2019good}, and instead use the power of a large language model~\citep{raffel2019exploring} that has a richer representation as it is created using more data.  
In this work, we translate language to a set of tokens that represent compositional value functions. To the best of our knowledge, our approach is novel, and maps language to a representation that is inherently compositional allowing us to solve novel tasks with few samples. 

Composing value functions was first demonstrated using the linearly-solvable MDP framework by \citet{todorov07}, in which value functions could be composed in an operator similar to disjunction to solve tasks~\citep{todorov09}.
This idea was extended to compose skills that achieve zero-shot disjunction~\citep{vanniekerk19} and approximate conjunction~\citep{haarnoja18,vanniekerk19,hunt19}.
More recently, \citet{tasse2020boolean} show that zero-shot optimal composition can be achieved for all three logical operators---disjunction, conjunction, and negation---in the stochastic shortest path setting. Our approach allows versatility in the type of expressions, and thereby the type of goal-based instructions that can be given to our agents.

\section{Conclusion}

In this work, we proposed an approach for instruction following that leverages the compositional representations present in both the Boolean Task algebra value functions and in large, pretrained language models.
Since regular value functions cannot in general be optimally combined to produce desired behaviours~\citep{todorov09, vanniekerk19}, we leveraged a recently introduced form of goal-oriented value functions that admit composability.
By ensuring that both the language and control aspects of the agent are compositional, we demonstrated that an agent can use its existing knowledge to quickly solve new tasks using very few samples.
Such sample efficiency is critical in developing long-lived agents that are required to learn and act in the real world.
In future work, we would like to create a fully differentiable model that learns to create a space of compositional goals and to map language to the space of Boolean algebra over the learned compostional goals.

\section* {Acknowledgements}

The authors acknowledge the Centre for High Performance Computing (CHPC) and the Mathematical Sciences Support Unit at the University of the Witwatersrand for providing computational resources for this work.

\newpage

\bibliography{main.bib}

\begin{appendices}

\section{Appendix A}

\begin{center}
\begin{tabular}{ cc }
\toprule
    Task primitive & Success rate\\
\midrule
    pickup\_ball & 0.997 $\pm$ 0.004\\
    pickup\_box & 0.996 $\pm$ 0.006\\
    pickup\_key & 1.000 $\pm$ 0.000\\
    pickup\_red & 0.996 $\pm$ 0.005\\
    pickup\_blue & 0.999 $\pm$ 0.003\\
    pickup\_green & 1.000 $\pm$ 0.000\\
    pickup\_grey & 0.996 $\pm$ 0.005\\
    pickup\_purple & 0.996 $\pm$ 0.005\\
    pickup\_yellow & 0.995 $\pm$ 0.008\\
\bottomrule
\end{tabular}
\captionof{table}{The mean success rate of the individual pretrained compositional value functions for each task primitive over 100 episodes. The standard deviations are over 10 runs.}
\label{policyperformance}
\end{center}

\begin{center}
\begin{tabular}{lc}
 \toprule
 \multicolumn{2}{c}{T5-small model parameters} \\
 \midrule
 Embedding dimension   & 512\\
 Fully-connected dimension &   2048\\
 Attention-heads & 8\\
 Encoder, Decoder Layers   & 6\\
 \bottomrule
\end{tabular}
\captionof{table}{The parameters of the T5-small model used in our experiments. To train the model we use the AdamW optimizer \cite{loshchilov2019decoupled} and a learning rate of 1e-4.}\label{t5params}
\end{center}

\section{Appendix B}
The DQNs used to learn the compositional value functions have the following architecture, with the CNN part being identical to that used by \citet{mnih15}:

\begin{enumerate}
    \item Three convolutional layers:
    \begin{enumerate}
        \item Layer 1 has 3 input channels, 32 output channels, a kernel size of 8 and a stride of 4.
        \item Layer 2 has 32 input channels, 64 output channels, a kernel size of 4 and a stride of 2.
        \item Layer 3 has 64 input channels, 64 output channels, a kernel size of 3 and a stride of 1.
    \end{enumerate}
    \item Two fully-connected linear layers:
        \begin{enumerate}
        \item Layer 1 has input size 3136 and output size 512 and uses a ReLU activation function.
        \item Layer 2 has input size 512 and output size 7 with no activation function.

    \end{enumerate}
\end{enumerate}

We used the ADAM optimiser with batch size 256 and a learning rate of $10^{-3}$.
The target Q-network was updated every 1000 steps, and we used $\epsilon$-greedy exploration, annealing $\epsilon$ from $1.0$ to $0.1$ over 1000000 timesteps.

\end{appendices}

\end{document}